\documentclass[10pt,twocolumn,letterpaper]{article}

\usepackage{3dv}
\usepackage{times}
\usepackage{epsfig}
\usepackage{graphicx}
\usepackage{amsmath}
\usepackage{amssymb}

\threedvfinalcopy 

\def\threedvPaperID{****} 

\ifthreedvfinal\fi
\begin{document}

\title{Plane-Based Optimization of Geometry and Texture for RGB-D Reconstruction of Indoor Scenes}

\author{Chao Wang and Xiaohu Guo\\
University of Texas at Dallas\\
800 W Campbell Rd, Richardson, TX 75080, USA\\
{\tt\small \{chao.wang3, xguo\}@utdallas.edu}
}

\maketitle

\begin{abstract}
   We present a novel approach to reconstruct RGB-D indoor scene with plane primitives. Our approach takes as input a RGB-D sequence and a dense coarse mesh reconstructed by some 3D reconstruction method on the sequence, and generate a lightweight, low-polygonal mesh with clear face textures and sharp  features without losing geometry details from the original scene. To achieve this, we firstly partition the input mesh with plane primitives, simplify it into a lightweight mesh next, then optimize plane parameters, camera poses and texture colors to maximize the photometric consistency across frames, and finally optimize mesh geometry to maximize consistency between geometry and planes. Compared to existing planar reconstruction methods which only cover large planar regions in the scene, our method builds the entire scene by adaptive planes without losing geometry details and preserves sharp features in the final mesh. We demonstrate the effectiveness of our approach by applying it onto several RGB-D scans and comparing it to other state-of-the-art reconstruction methods. 
\end{abstract}

\section{Introduction}

Online and offline RGB-D reconstruction techniques are developing fast in recent years with the prevalence of consumer depth cameras, and they have wide applications in many popular areas such as virtual reality (VR), augmented reality (AR), gaming and 3D printing. State-of-the-art online 3D reconstruction methods can capture indoor and outdoor scenes in the real-world environments efficiently with geometry details \cite{niessner2013real,whelan2015elasticfusion,whelan2016elasticfusion,dai2017bundlefusion,prisacariu2017infinitam}. However, resulting 3D models of these methods are usually too dense with unsatisfying textures due to many reasons including noisy depth data, incorrect camera poses and oversmoothing in data fusion. These models can not be used directly in most applications without further refinement or post-processing.

In order to lower the density of reconstructed models and improve the structure quality, one typical strategy is to introduce plane primitives into front-end (such as tracking) or back-end (such as model refinement) of reconstruction pipeline, as typical indoor scenes are primarily composed of planar regions, especially buildings and houses with structure following Manhattan-world assumption. However, almost all existing planar reconstruction methods take into account only large planar regions such as walls, ceilings, floors and large table surfaces, and simply ignore and remove other objects with freeform surfaces no matter if they contain planar regions or not, such as various indoor furniture and objects on or adhering to large planes. These reconstructed models with only large planes are too simplified and lack fidelity that they are not applicable to many situations acquiring geometry details such as indoor roaming with decorations, gaming and various VR or AR applications. Moreover, one challenging problem in planar reconstruction of indoor scene is that geometry details are usually noisy because of noisy RGB-D raw data, and it is difficult and also time-consuming to extract plane primitives or other types of geometry priors from the scene while still preserving the original shape. 

In this paper, we present a novel approach to reconstruct RGB-D indoor scene using planes and generate a lightweight and complete 3D textured model without losing geometry details. Our method takes as input a RGB-D sequence of indoor scene and a dense coarse mesh reconstructed by some online reconstruction method on this sequence. We firstly partition the entire dense mesh into different planar clusters, each of which represents a plane primitive. Next, we simplify the dense mesh based on the planar partitions into a lightweight mesh which preserves both large planar regions as well as small objects without losing geometry details. In order to generate texture mapping for mesh faces, we create texture patch for each plane and sample points on the plane, and run a global optimization process to maximize the photometric consistency of sampled points across frames by optimizing camera poses, plane parameters and texture colors. Finally, we optimize the mesh geometry by maximizing consistency between geometry and plane primitives, which further preserves sharp features of original scene such as edges and corners of plane intersections.

Our approach can be regarded as a back-end RGB-D reconstruction framework after online reconstruction process. Experiments show that our method exceeds state-of-the-art RGB-D planar reconstruction method in keeping geometry details and sharp features in the result lightweight 3D textured models.

\section{Related work}
\label{sec:related}

Online and offline 3D reconstruction with consumer depth cameras have been widely studied in recent years. One major category of online large-scale 3D reconstruction methods utilizes volume data structure and volumetric fusion technique, like one of the earliest online method KinectFusion \cite{Newcombe01} and its subsequent variants such as scalable online reconstruction \cite{chen2013scalable}, VoxelHashing \cite{niessner2013real}, Volume-shifting \cite{whelan2015real}, BundleFusion \cite{dai2017bundlefusion} and InfiniTAM \cite{prisacariu2017infinitam}. One important advantage of these volume-based methods is that it is fast and efficient to smooth noisy data scanned by depth cameras and easy to generate mesh with connected surface. However, a major limitation is that volumetric fusion oversmoothes the surface and removes the sharp features since it tends to average points and their colors in the same voxel position in the volume. Besides volume-based method, another strategy is point-based method, which utilizes surfels as basic data structure to represent the surface without connectivity \cite{keller2013real,whelan2015elasticfusion,whelan2016elasticfusion}. However, it still suffers from oversmoothing problem since it also tends to make a moving average of points nearby in the space \cite{whelan2015elasticfusion}. Moreover, error accumulates in surface representation and camera poses through the scanning process because of oversmoothing as well as other factors including noisy color and depth data.

In order to improve camera pose estimation and also the quality of surface, several works introduce planes into the reconstruction pipeline to estimate complicated surface with simple surface priors, especially in buildings or indoor scene with special structure features like Manhattan-world assumption. One category of methods introduce plane constraints into front-end of the reconstruction process to improve robustness of camera tracking. One of the earliest methods by Dou et al. \cite{dou2012exploring} combines both plane and feature point correspondences to estimate camera poses in RGB-D scanning. A recent work by Hsiao et al. \cite{hsiao2017keyframe} uses very similar idea to introduce plane constraint into tracking in SLAM framework, and their result exceeds current state-of-the-art online 3D reconstruction methods in pose estimation. For offline reconstruction framework, a very recent work by Halber and Funkhouser \cite{halber2017fine} proposes a fine-to-coarse global registration algorithm on RGB-D data by combining planar relationship constraints with other types of constraints, and their method works efficiently on large-scale RGB-D data. 

Another category of methods tries to utilize planes to better represent the surfaces of final model reconstructed from RGB-D data. Dzitsiuk et al. \cite{dzitsiuk2017noising} introduce plane priors in real-time 3D reconstruction pipeline on mobile devices to lower the computation complexity, de-noise depth data and improve indoor structure. However, this method only refines several limited types of planes like walls, ceilings and floors and leaves other scanned data unchanged. Another impressive method 3DLite \cite{huang20173dlite} is the latest method we found to reconstruct RGB-D model by planes and optimize face textures on them. This method generates lightweight model with optimized textures from an input dense and fused reconstruction. It proposes a frame-based plane detection technique to extract large planes from RGB-D mesh, and then globally optimizes camera poses and face textures on the planes. However, this method still cannot detect and simply removes all small planes and non-planar geometry details, whereas our method is to reconstruct the entire scene by planes without losing geometry details.

Planar reconstruction from point clouds is also a hot topic for decades, as point cloud data is prevalent and can be easily acquired via various tools such as Laser scanners or structure from motion (SfM) data. Planes are widely used in reconstruction of outdoor environments to fit building structure. Monszpart et al. \cite{monszpart2015rapter} extract regular arrangement of planes from unstructured, large-scale and noisy building scans, and reconstruct the building model with complete and lightweight planes. However, it only detects large planar regions in the building framework and ignore details inside, so it is more like an architecture reconstruction method using planes. So is another method proposed by Mura et al. \cite{mura2016piecewise} that reconstructs only the permanent components (walls, ceilings and floors) of buildings by extracting planes on these components from all detected ones from input point cloud. 

The two most common plane primitive detection strategies are RANSAC and region growing. RANSAC-based method is popular for its simplicity, scalability and probabilistic guarantees. However, it easily misses global scene-level structures and ignores the neighborhood regularity between points \cite{monszpart2015rapter}.  Therefore, RANSAC-based method are usually used on regular models such as CAD shapes and building structure \cite{li2011globfit}. Compared with RANSAC, region growing expands from seeds to neighbors and inherently detects parts that are connected, and is more suitable and widely used in plane detection of large-scale models. However, one important disadvantage is that region growing is not designed to detect planes on free-form shapes since it easily partitions curved surfaces into irregular parts and causes over-segmentation. Chauve et al. \cite{chauve2010robust} propose one of the earliest methods to extract planes and build triangular mesh from noisy unstructured point cloud with planar regions. Boulch et al. \cite{boulch2014piecewise} present a 3D reconstruction of a piecewise-planar surface from range images by regularizing the surface with respect to edges and corners. Oesau et al. \cite{oesau2016planar} propose a planar shape detection and regularization method in tandem from raw point sets. Similarly, even though these methods do not ignore non-planar regions on purpose,
they rely on region growing and still cannot preserve shape of curved surfaces.

\section{Planar reconstruction pipeline}
\label{sec:pipeline}

Our reconstruction pipeline takes a RGB-D sequence as input, and firstly uses some state-of-the-art online reconstruction such as BundleFusion \cite{dai2017bundlefusion} to reconstruct an initial dense mesh. Initially, we detect plane primitives by partitioning the mesh into different clusters, each of which represents a plane patch (Section \ref{sec:partition}). The following step is to simplify the dense mesh based on the planar partition (Section \ref{sec:simp}). Next, we sample points on each plane in the mesh and create texture patch for the plane patches, and optimize camera poses, plane parameters and texture colors to maximize the color consistency across frames (Section \ref{sec:opt:plane}). Finally, we optimize the mesh geometry to maximize consistency between vertices on the mesh and corresponding plane primitives (Section \ref{sec:opt:geo}).

\subsection{Mesh planar partition}
\label{sec:partition}

Unlike existing planar reconstruction methods which only take into account large planar regions, we aim to partition the entire mesh into plane primitives to include all geometry details. As we introduce in Section \ref{sec:related}, two most common plane detection method is RANSAC and region growing. RANSAC-based detection is efficient but easily misses global scene-level structures, while region growing is more robust in detecting planes under regularity but still is not designed to detect planes in freeform shapes with curved surfaces. 3DLite \cite{huang20173dlite} proposes a frame-based plane detection method to extract large planes from RGB-D mesh. However, this method still cannot detect small planar regions or non-planar surfaces and it simply removes them in the final mesh. Moreover, the frame-by-frame plane detection and merging scheme used in 3DLite is time-consuming and easy to accumulate errors so it has to utilize many further optimization steps for robustness.

\begin{figure}[!]
\begin{center}
\includegraphics[width=0.5\textwidth]{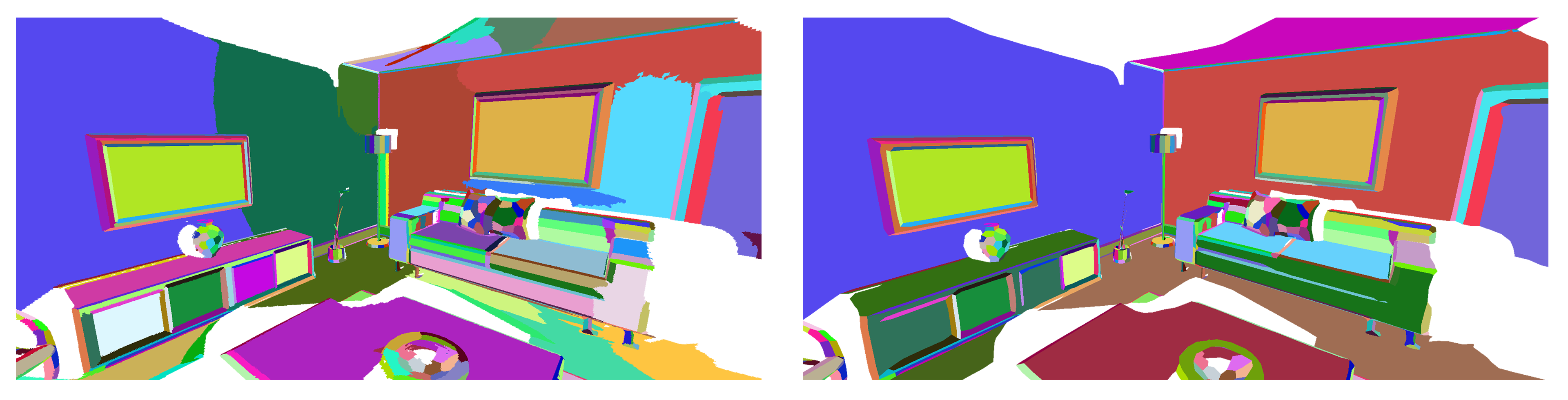}
\end{center}
\caption{Plane partition result on a mesh without plane merging (left) and with merging (right) on the `lr kt2n' model from ICL-NUIM dataset. Notes that small neighbor plane patches on the walls, floors and ceilings are merged into large planes.}
\label{fig:partition}
\end{figure}

In our approach we refer to a state-of-the-art surface partition algorithm proposed by Cai et al. \cite{cai2017surface}. This method proposes a new principle component analysis (PCA) based energy, whose minimization leads to an optimal piecewise-linear planar approximation of the entire surface with high quality. After an input mesh is partitioned into clusters, each cluster is attached with a plane proxy $\mathbf{\phi}_i$ defined by the centroid $\mathbf{c}_i$ and normal $\mathbf{n}_i$ as the smallest eigenvector direction from the covariance matrix of the cluster. In order to detect planes more sensitively and regularize the compactness of planar clusters, in our experiments we lower the coefficient $\alpha$ used in Eq. (4) of the paper \cite{cai2017surface} from $10^{-15}$ to $10^{-20}$. Figure \ref{fig:partition} (left) shows planar partition result on a dense mesh reconstructed by online reconstruction system VoxelHashing \cite{niessner2013real} with groundtruth camera poses.

After we get the initial planar partition, we run a further plane merging step to merge \textbf{adjacent} planes together into large ones (Figure \ref{fig:partition} right). The reason is that, the reconstructed mesh on noisy RGB-D data always contains noise such as bumpy points on planar regions. Because of this, a large planar region is possibly partitioned into several different small clusters instead of only one large cluster (see the walls in Figure \ref{fig:partition} left). Our plane merging criteria is like this: two adjacent planes $i$ and $j$ can be merged together if they satisfy all following conditions:
\begin{enumerate}
\item[(1)] Angle between two plane normals are small:  
$\theta(\mathbf{n}_i, \mathbf{n}_j) < \epsilon_{\mathbf{n}}$;
\item[(2)] Average distance between two planes is small: $Avgdis(i, j) < \epsilon_d, \ \ Avgdis(j, i) < \epsilon_d$;
\item[(3)] Angle between the ray connecting two plane centers $\mathbf{r}_{ij} = \mathbf{c}_i - \mathbf{c}_j$ and each plane normal is as close as possible to $90^{\circ}$:
$|\cos(\theta(\mathbf{r}_{ij}, \mathbf{n}_i))| < \epsilon_{\mathbf{c}}, \ \ |\cos(\theta(\mathbf{r}_{ij}, \mathbf{n}_j))| < \epsilon_{\mathbf{c}}$.
\end{enumerate}
Here $\theta(\cdot, \cdot)$ is the angle between two vectors, and $Avgdis(i, j)$ is the average distance between all vertices in cluster $i$ and plane $j$. In our experiment we use $\epsilon_{\mathbf{n}} = 8^{\circ}, \epsilon_d = 0.05, \epsilon_{\mathbf{c}} = \cos(80^{\circ})$. The first two conditions are from the plane merging method by Dzitsiuk et al. \cite{dzitsiuk2017noising}. We add the third rule to get rid of special cases that two adjacent planes are almost `overlapped' in the noisy mesh, and only merge neighbor planes that are on one side of each other.

\subsection{Mesh simplification}
\label{sec:simp}

After partitioning the mesh into plane clusters, we simplify the mesh based on clusters to create a lightweight mesh for further optimization. Note that even though we already have a model composed of planes, we still choose to create a mesh by simplifying the original dense mesh instead of using some mesh generation algorithm (such as Delaunay triangulation) on planes which strategy appears in most existing methods \cite{chauve2010robust, huang20173dlite, li2011globfit}. The reason is that, we found that it is difficult and also time-consuming to create correct connectivity from complicated plane interceptions in a noisy model, especially an indoor reconstruction mesh containing various geometry objects with free-form shapes. Therefore, our strategy is to efficiently create an initial lightweight mesh through simplifying the original dense mesh based on planes, and further optimize its geometry to fit the planes (Section \ref{sec:opt:geo}).

\begin{figure}[h]
\begin{center}
\includegraphics[width=0.49\textwidth]{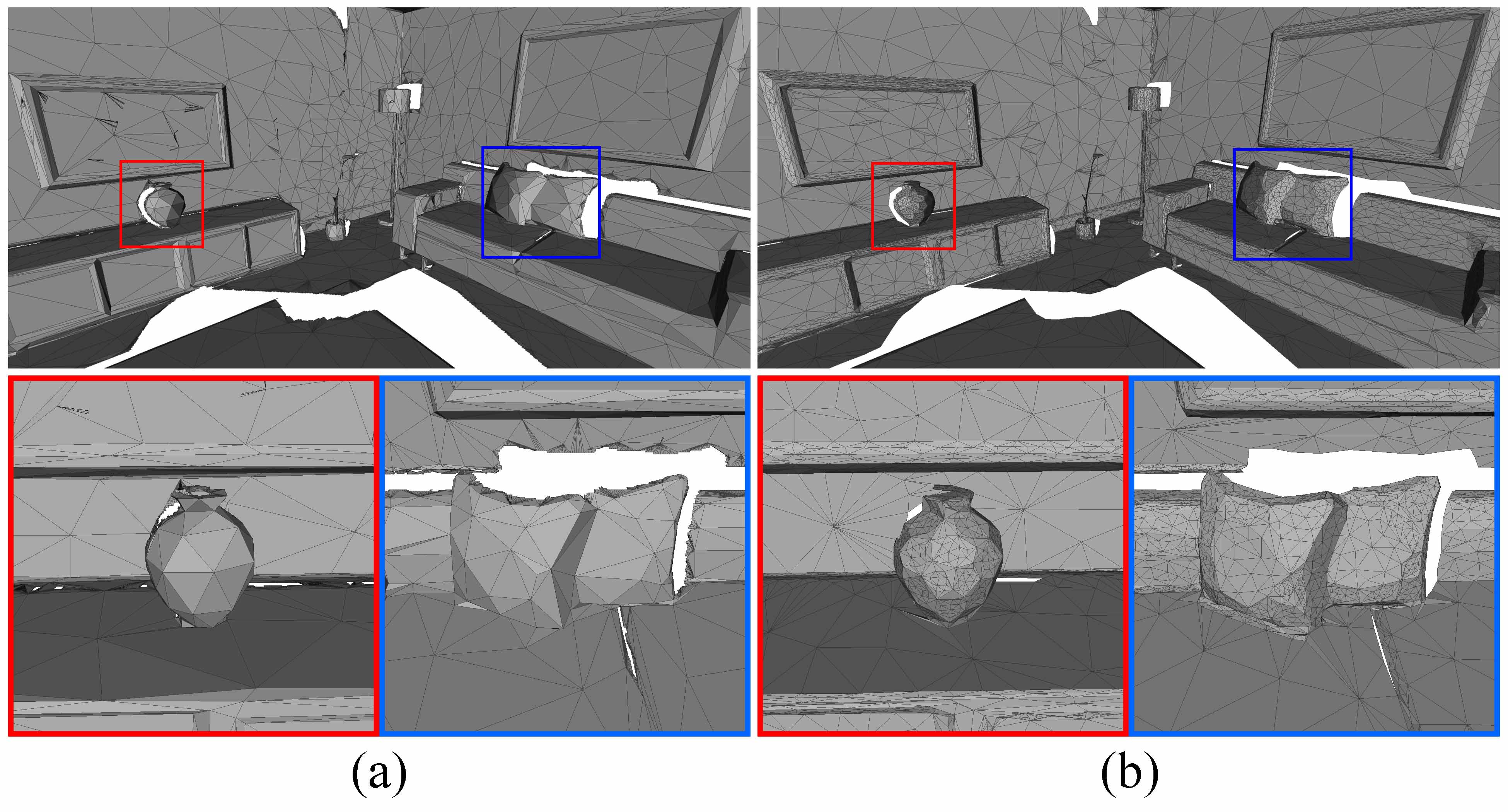}
\end{center}
\caption{Simplified mesh by standard QEM (a) and our method (b) on the scan `lr kt2n' from ICL-NUIM dataset. Note for sharp features such as plane boundaries, and curved surfaces such as vase and pillows.}
\label{fig:simp}
\end{figure}

During mesh simplification, we choose the classic quadric error metric (QEM) based surface simplification \cite{garland1999quadric}. The core of QEM is to contract edges by minimizing a point-to-plane energy, which suits our purpose to preserve plane shapes very well. Moreover, QEM is simple and efficient and also tends to preserve sharp features of original model, which is also what we need in the simplified mesh. The difference is that, unlike a standard QEM process which puts all edges into a minimum heap based on contraction energy and contracts them globally, our simplification process works cluster-by-cluster and is composed of two steps: (1) fix all cluster boundary edges and simplify only inner-cluster edges of each cluster separately; (2) fix simplified inner-cluster edges and simplify all boundary edges of clusters. Through this two-step process, we can reduce unimportant information inside planes while still preserving important sharp features which usually appear along plane boundaries. Another consideration is that cluster sizes are different from each other, but we prefer every cluster contains almost the same number of edges after simplification. That is, large planes contain large triangles and vise versa. Then, simplifying cluster-by-cluster with a same contraction target is better to make small clusters denser to preserve their shapes, as they are usually the partition of curved surfaces. Figure \ref{fig:simp} shows comparison between standard QEM (keep mesh boundary) and our method on the same model used in Figure \ref{fig:partition}. The two simplified models have exactly the same number of faces 41K while mesh QEM model has 30K points, larger than our model with 22K points. Obviously QEM oversimplifies plane boundaries and curved surfaces while our method keeps the sharp features and geometry details better.

\subsection{Plane, camera pose and texture optimization}
\label{sec:opt:plane}

After we get the simplified mesh with plane partition, we run an optimization process to maximize the photometric consistency for the mesh geometry between frames. Before optimization, we still need to do some pre-processing on the mesh.

The first work is to generate an initial texture mapping for all the faces of the mesh. In our method we create a 2D texture patch for each 3D plane on the mesh. Even though there are many mature parameterization methods to generate 2D texture patches for 3D mesh surface, we use a very simple and efficient one. Considering that vertices in each cluster on the mesh are already near co-planar, we simply project them onto the plane to get the 2D patch, and then sample grid points inside the patch boundary to get texel points. In our experiment we use a fixed sampling density 0.0025m. That is, 1 meter in global space corresponds to $1.0/0.0025 = 400$ pixels in the texture image. Clearly, each texel is located inside some projected triangluar face from the mesh. Then, we compute each texel's barycentric coordinates inside its corresponding 2D face, and use them to compute the texel's corresponding 3D point $\mathbf{p}$ in global space by simply applying the same barycentric coordinates onto the three vertices of the face in 3D space. These 3D texel points will be used as the major elements during optimization process.

Another thing to mention is about the keyframes selected from RGB-D frames. To reduce time complexity and increase texture quality, we follow the similar idea of color map optimization by Zhou and Koltun \cite{zhou2014color} to select only sharp frames in every interval. Similar to their method, we quatify the blurriness of each image with the metric by Crete et al. \cite{crete2007blur}. The difference is that, instead of selecting keyframe for every 1 to 5 seconds in \cite{zhou2014color}, we simply select the sharpest frame in every 5 or 10 frames depending on the data.

The input in our optimization process is color images $\{\mathbf{I}_i\}$ and depth images of keyframes, all texels' 3D points $\{\mathbf{p}\}$ sampled on the mesh, initial camera poses $\mathbf{T} = \{\mathbf{T}_i\}$ (global to camera space) and initial plane parameters $\Phi = \{\mathbf{\phi}_j\}$. During the optimization, we maximize the photo consistency of 3D texels' projections on corresponding planes across frames by optimizing camera poses, plane parameters and texture colors. More specifically, our objective function is 
\begin{equation}
E_{tex}(\mathbf{T}, \Phi, \mathbf{C}, \mathbf{F}) = E_{c}(\mathbf{T}, \Phi, \mathbf{C}, \mathbf{F}) + \lambda_1 E_{p}(\Phi) + \lambda_2 E_{s}(\mathbf{F}),
\label{eq:main}
\end{equation}
where $E_{c}$ is photometric consistency energy, $E_{p}$ is constraint for planes, $E_{s}$ is constraint for non-rigid correction offsets for color images introduced in \cite{zhou2014color}, and $\lambda_1$ and $\lambda_2$ are coefficients to balance terms. In our experiment we use $\lambda_1$ such that initial values of $E_{c} = \lambda_1E_{p}$ and $\lambda_2 = 0.1$.

\textbf{Photometric consistency term.} The photometric energy is designed to measure the photometric error between color of each texel's projection point on its corresponding plane and its target color across frames:
\begin{equation}
E_{c}(\mathbf{T}, \Phi, \mathbf{C}, \mathbf{F}) = \sum_i \sum_{\mathbf{p} \in \mathbf{P}_i} ||C(\mathbf{p}) - \mathbf{I}_i(\mathbf{F}_i(\pi(\mathbf{T}_i\mathbf{q})))||^2,
\label{eq:color}
\end{equation}
where $C(\mathbf{p})$ is the target color for $\mathbf{p}$, and $\mathbf{P}_i$ is set of all visible 3D texels in frame $i$, and $\pi$ is the perspective projection on homogeneous coordinate $\mathbf{v} = (v_0, v_1, v_2, v_3)^\top$: 
\begin{equation}
\pi(\mathbf{v}) = (\frac{v_0 * f_x}{v_2} + c_x, \frac{v_1 * f_y}{v_2} + c_y)^\top
\end{equation}
where $c_x, c_y, f_x, f_y$ are principal point and focal lengths from camera intrinsic matrix, respectively. $\mathbf{F} = \{\mathbf{f}_{i,l}\}$ in Eq. (\ref{eq:color}) is the set of non-rigid correction fuctions of control vertices over color image $\mathbf{I}_i$ \cite{zhou2014color}:
\begin{equation}
\mathbf{F}_i(\mathbf{u}) = \mathbf{u} + \sum_{l}\delta_l(\mathbf{u})\mathbf{f}_{i,l},
\end{equation}
where $\delta_l$ is the basis functions for bilinear interpolation, $\mathbf{f}_{i,l}$ is 2D correction vector for $l$th control vertex in the orthogonal grid in color image of frame $i$. In our experiment we follow the same parameters from \cite{zhou2014color} to use $20 \times 16$ grid on each image. $\mathbf{q}$ in Eq. (\ref{eq:color}) is the projection point from $\mathbf{p}$ onto its corresponding plane $\phi(\mathbf{p})$ represented by 3D normal $\mathbf{n}_{\mathbf{p}}$ and a scalar $w_{\mathbf{p}}$:
\begin{equation}
\mathbf{q} = \mathbf{p} - (\mathbf{p}^\top \mathbf{n}_{\mathbf{p}} + w_{\mathbf{p}})\mathbf{n}_{\mathbf{p}},
\label{eq:proj}
\end{equation}

\textbf{Plane constraint term.} Plane constraint term is to minimize the sum of distances from 3D texel points to their corresponding planes:
\begin{equation}
E_{p}(\Phi) = \sum_{\mathbf{p}}||\mathbf{p}^\top \mathbf{n}_{\mathbf{p}} + w_{\mathbf{p}}||^2
\end{equation}

\textbf{Offset constraint term.} The offset constraint is to minimize the magnitude of offset vectors:
\begin{equation}
E_{s}(\mathbf{F}) = \sum_i\sum_l\mathbf{f}_{i,l}^T\mathbf{f}_{i,l}
\end{equation}

\begin{figure}[h]
\begin{center}
\includegraphics[width=0.5\textwidth]{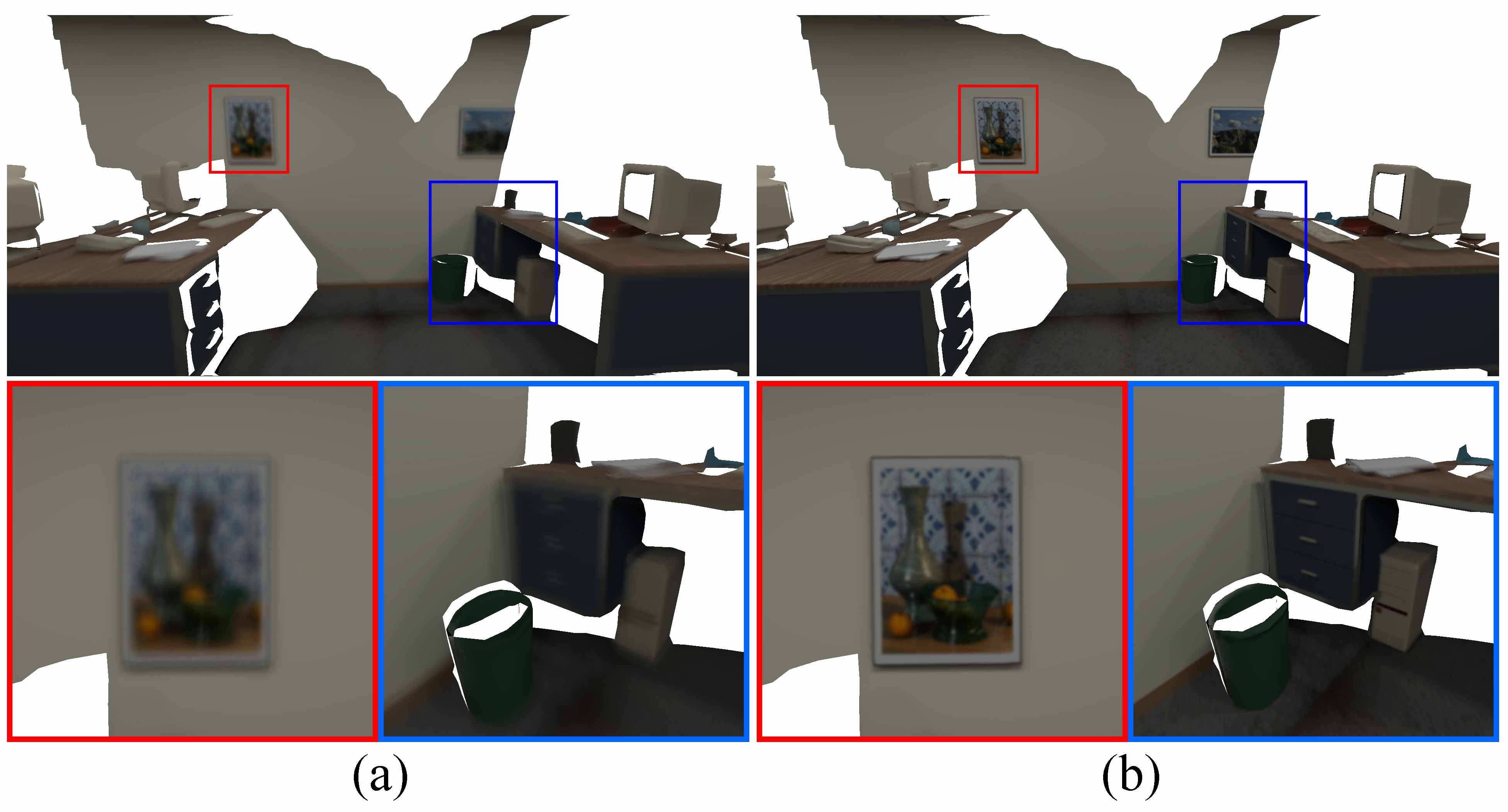}
\end{center}
\caption{Mesh with textures before and after texture optimization on the simplified model of sequence `of kt2' from ICL-NUIM dataset \cite{handa:etal:ICRA2014}. (a) Texture before optimization. Each texel's color is the average color from all visible color frames. (b) Texture after optimization.}
\label{fig:texture}
\end{figure}

In order to minimize the objective function in Eq. (\ref{eq:main}), we follow the similar alternating optimization strategy in \cite{zhou2014color}. The basic idea is to alternate between optimizing different variables with some others fixed. In each iteration we firstly optimize $\mathbf{C}$ and fix all the others, and next optimize $\Phi$ and fix the others, and finally optimize $\mathbf{T}$ and $\mathbf{F}$ and fix the others. When optimizing $\mathbf{C}$, the problem becomes a linear system with closed form solution that $C(\mathbf{p})$ is the average color of all projected pixels associated with $\mathbf{p}$ in all visible frames. When optimizing $\Phi$, we use standard Gauss-Newton method to update the 4 parameters of each plane directly. Note that the optimization of each plane is independent with others so we can solve them in parallel. Optimizing $\mathbf{F}$ is similar as optimizing $\Phi$ that we update all relevant variables directly. When optimizing $\mathbf{T}$, we parameter each pose $\mathbf{T}_i$ by 6 parameters (3 for rotation, 3 for translation) as the incremental transformation, and locally linearize each pose around its last updated value. Similarly, optimization of each $\mathbf{T}_i$ and $\mathbf{F}_i$ are also independent with other frames and can be solved in parallel. Figure \ref{fig:texture} shows textures on a mesh before and after our optimization process. Before optimization, we use average color from all visible frames for each texel. Our optimization process can reduce noise and make texture clearer.

\subsection{Geometry optimization}
\label{sec:opt:geo}

The final step is to optimize the mesh geometry to fit the planes as close as possible. The fused mesh reconstructed from RGB-D data always contain noise or oversmoothed surfaces, such as bumpy surfaces on planar regions and smoothed borders which suppose to be sharp features. By optimizing mesh geometry to fit the optimized planes, we can reduce noise from mesh surface and sharpen geometry features.

In order to optimize the consistency between geometry and planes, our method is to maximize the consistency between mesh vertices in each cluster and their corresponding planes. As we introduced in Section \ref{sec:opt:plane}, each 2D texel is located inside a triangular face's projection. During the optimization, we utilize the initial barycentric relationship between each texel and its corresponding face, and try to preserve this relationship between texel points' projections on planes and the optimized vertices in each face. Specifically, we want to minimize the following energy function w.r.t. all vertices $\mathbf{V} = \{\mathbf{v}_i\}$:
\begin{equation}
E_{vert}(\mathbf{V}) = E_g(\mathbf{V}) + \lambda_3 E_t(\mathbf{V}).
\label{eq:geo}
\end{equation}
Here $E_g$ is the geometry consistency term
\begin{equation}
E_g(\mathbf{V}) = \sum_{\mathbf{p}}||\mathbf{q} - \sum_{i=0}^2 b_{\mathbf{p},i}\mathbf{v}_{f_{\mathbf{p}}, i}||^2,
\end{equation}
where $\mathbf{q}$ is the projection from 3D texel point $\mathbf{p}$ onto its corresponding plane as described in Eq. (\ref{eq:proj}), $f_{\mathbf{p}}$ is index of the face $\mathbf{p}$ corresponds to, $\mathbf{v}_{f_{\mathbf{p}}, i}$ is the $i$th vertex of face $f_{\mathbf{p}}$, and $b_{\mathbf{p},i}$ is $\mathbf{p}$'s barycentric coordinate corresponding to the $i$th vertex in face $f_{\mathbf{p}}$. 

Another term $E_t$ in Eq. (\ref{eq:geo}) is a regularization term to minimize the difference between each vertex and the mass center of all its neighbors:
\begin{equation}
E_t(\mathbf{V}) = ||\mathbf{LX}||_F^2.
\end{equation}
Here $\mathbf{X} = [\mathbf{v}_1, \mathbf{v}_2, \cdots, \mathbf{v}_n]^\top$ is matrix of target vertices we want to compute, with $n$ the number of vertices on the mesh. $\mathbf{L}$ is $n\times n$ matrix denoting the discrete graph Laplacian matrix based on the connectivity of the mesh, and its elements are
\begin{equation}
\mathbf{L}_{ii} = 1,\ \ \ \mathbf{L}_{ij}=
\begin{cases} 
-\frac{1}{|N(i)|} & j \in N(i)\\
0 & j \notin N(i)
\end{cases},
\end{equation}
where $N(i)$ is the set of $\mathbf{v}_i$'s neighbor vertices on the mesh. That is, we want to minimize the difference between each optimized vertex and the average of its neighbor vertices. The term $E_t$ is added to ensure that problem in Eq. (\ref{eq:geo}) has valid solutions. $\lambda_3$ in Eq. (\ref{eq:geo}) is for balancing the two terms and we simply use $\lambda_3 = 1.0$.

\begin{figure}[h!]
\begin{center}
\includegraphics[width=0.5\textwidth]{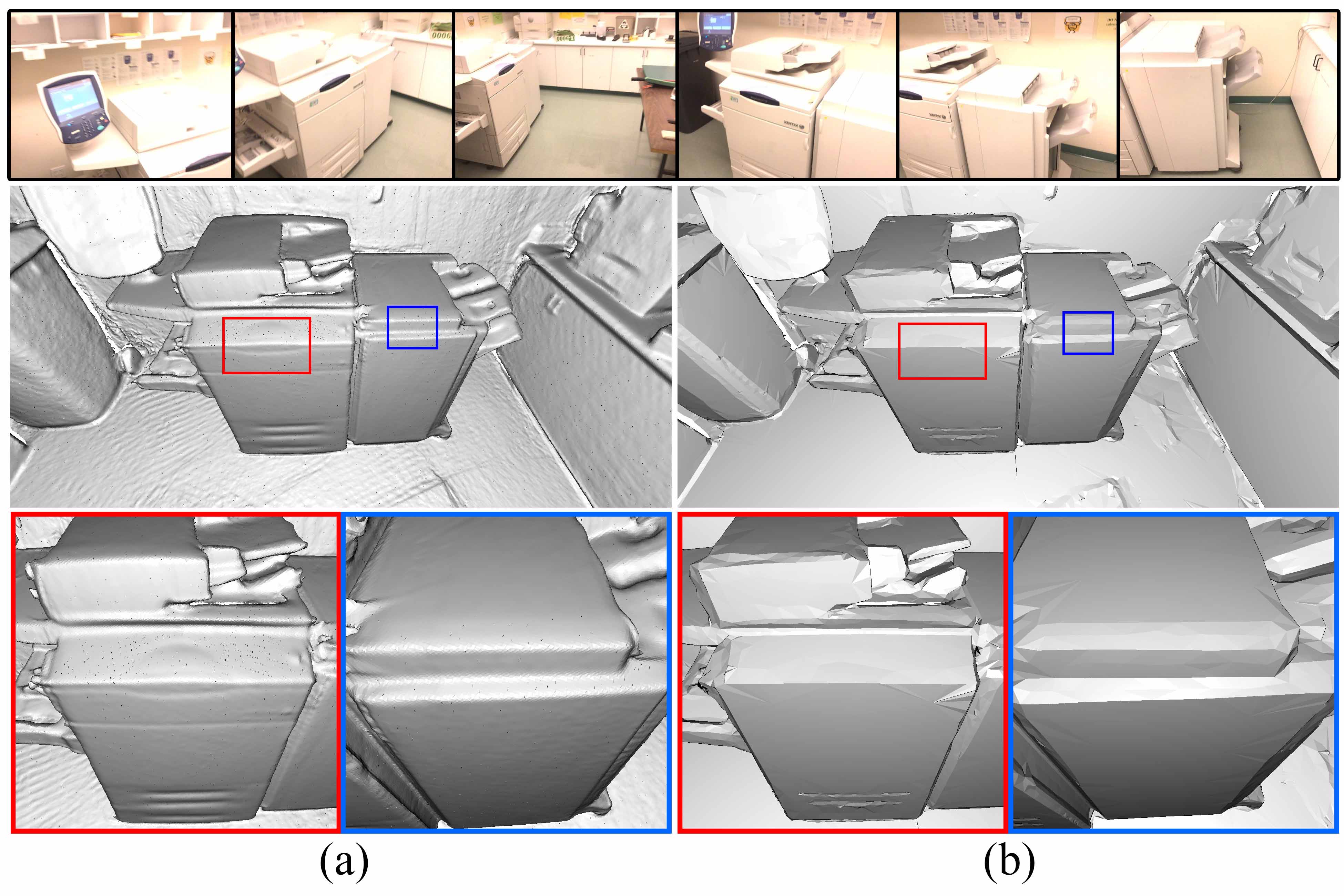}
\end{center}
\caption{Comparison between reconstructed mesh with or without geometry optimization on model `copyroom' from BundleFusion dataset \cite{dai2017bundlefusion}. First row shows selected input color images of the scene. Note for the sharp edges between planes on the printer. (a) Fused dense mesh from BundleFusion system. Note that sharp features are oversmoothed. (b) Our simplified mesh after geometry optimization. Note that sharp features are clear.}
\label{fig:sharp}
\end{figure}

The problem in Eq. (\ref{eq:geo}) is actually a sparse linear system and can be solved by Cholesky decomposition efficiently. Figure \ref{fig:sharp} shows result mesh with optimized geometry on a scan from BundleFusion dataset \cite{dai2017bundlefusion}. Even though the input dense model is oversmoothed on sharp feature places, our method can preserve the sharp features in the final lightweight mesh.

\section{Results}

\begin{figure}[h]
\begin{center}
\includegraphics[width=0.48\textwidth]{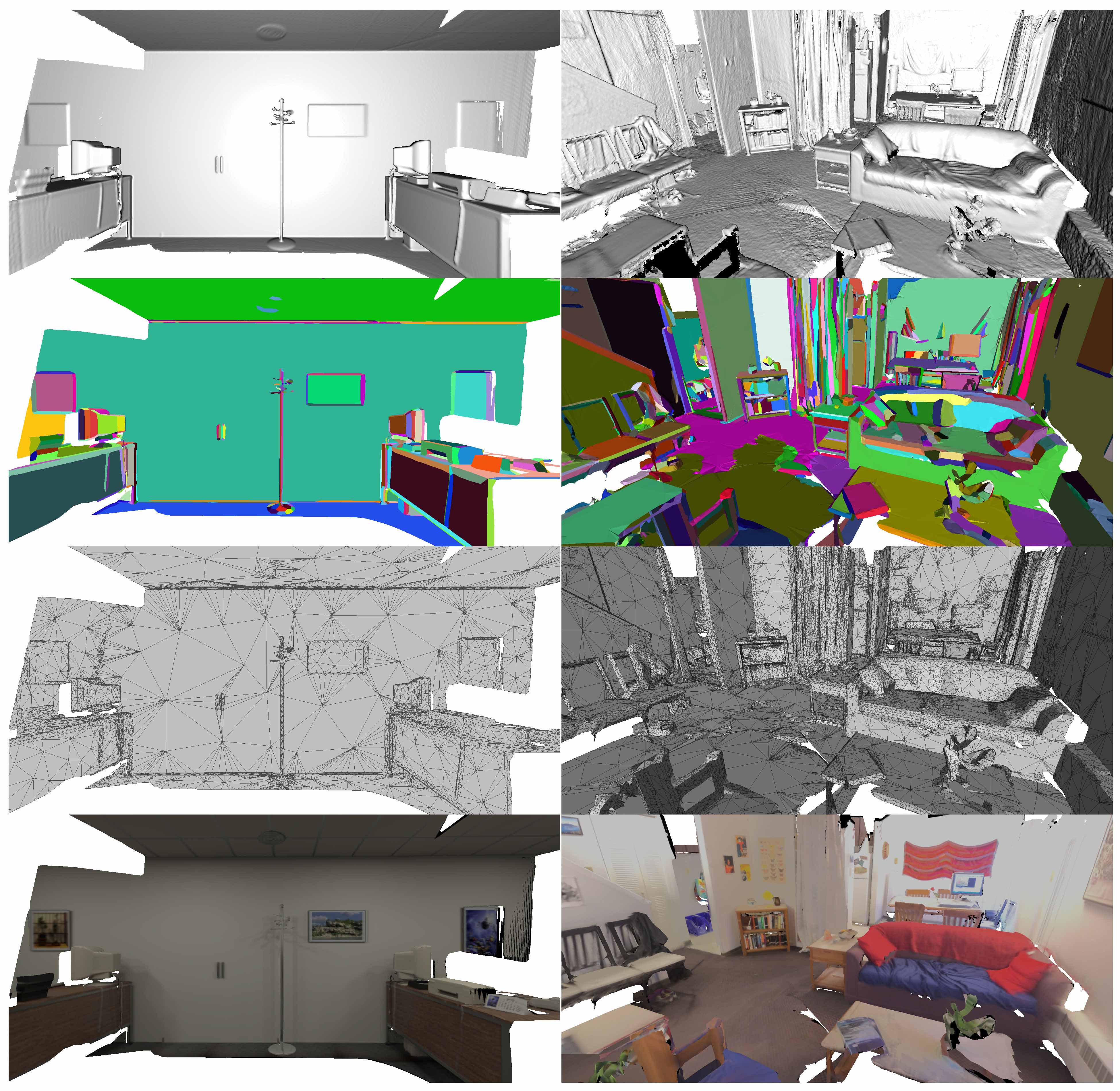}
\end{center}
\caption{Our result on two RGB-D scans: `of kt2' (left column) from ICL-NUIM dataset \cite{handa:etal:ICRA2014} and `apt0' (right column) from BundleFusion dataset \cite{dai2017bundlefusion}. First row: input dense model. Second row: plane partition. Third row: result mesh. Forth row: mesh with textures.}
\label{fig:ours1}
\end{figure}

\begin{figure*}[!]
\begin{center}
\includegraphics[width=1.0\textwidth]{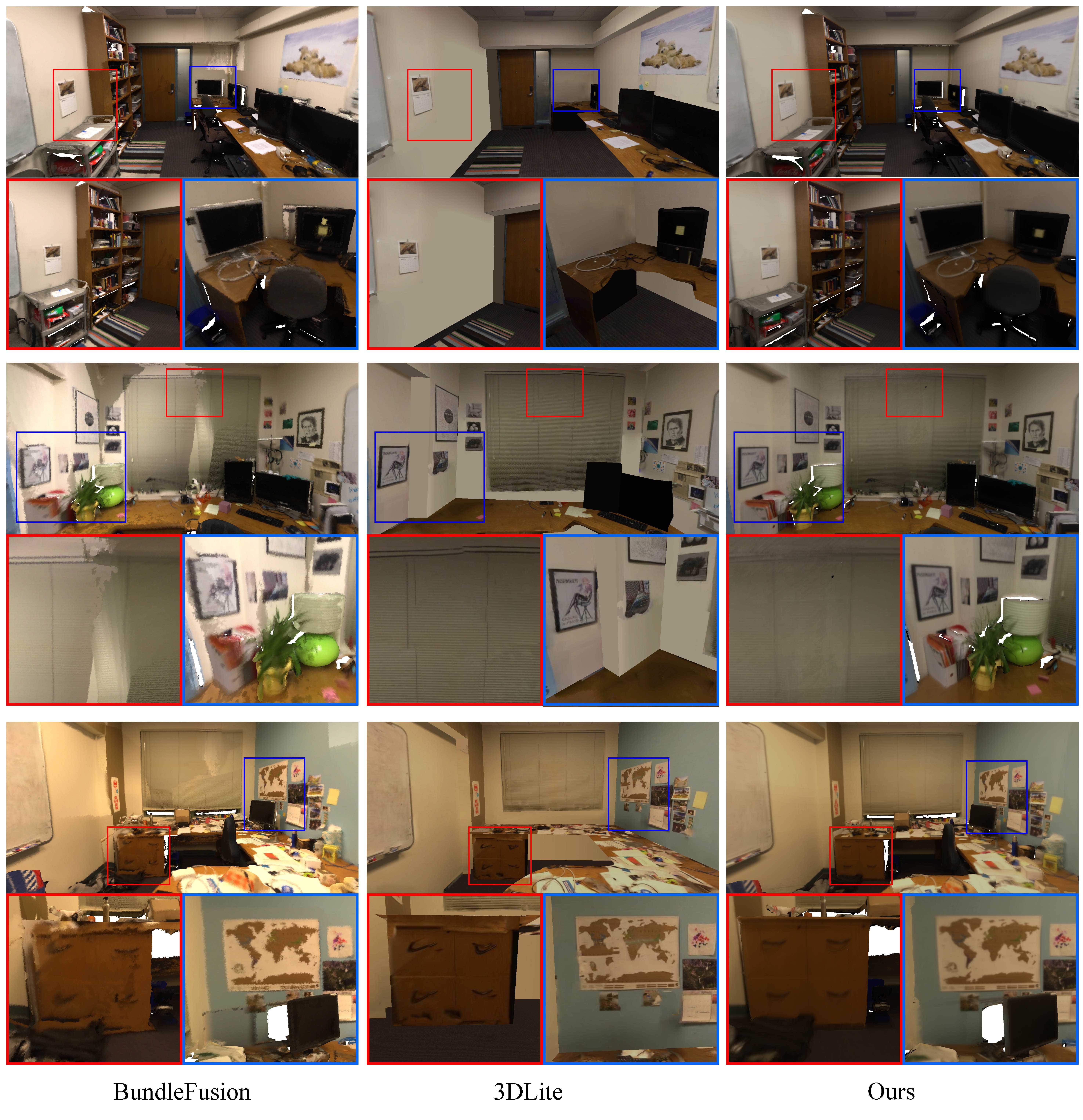}
\end{center}
\caption{Result comparison between BundleFusion \cite{dai2017bundlefusion}, 3DLite \cite{huang20173dlite} and ours on two scans `office0' (first two rows) and `office3' (third row) both from BundleFusion dataset. The color texture in BundleFusion models is default vertex colors fused from the reconstruction system, while others are optimized face texture. }
\label{fig:comp}
\end{figure*}

We tested our method on 10 scans from three popular RGB-D datasets: 6 models from BundleFusion \cite{dai2017bundlefusion} (the first 6 rows in Table \ref{tables}), 2 from ICL-NUIM \cite{handa:etal:ICRA2014} (the following 2 rows in Table \ref{tables}) and 2 from TUM RGB-D dataset \cite{sturm12iros} (the last 2 rows in Table \ref{tables}). According to Huang et al. \cite{huang20173dlite}, the online BundleFusion code possibly generates some artifacts in the resulting reconstruction. So in our experiment, we follow the same idea of \cite{huang20173dlite} to run VoxelHashing code \cite{niessner2013real} on each RGB-D sequence to reconstruct our input dense mesh using groundtruth poses (BundleFusion dataset provides poses computed by BundleFusion system). Table \ref{tables} shows quantitative results of each scan and our result models. Note that the number of faces or vertices of each result model is only 1\%-3\% of that of original dense model. Figure \ref{fig:ours1} and \ref{fig:comp} show more qualitative results of two scans. These figures show that our method can generate a lightweight mesh with preserved sharp features and good face textures.

For time analysis, we implemented our method in C++ and tested on a desktop with Intel Core i7 2.5GHz CPU and 16 GB memory. The running time on each scan is shown in Table \ref{tables}. The time data in the table is total time of our complete pipeline, of which plane partition (Section \ref{sec:partition}) and mesh simplification (Section \ref{sec:simp}) steps both take about 2 min on each model, while geometry optimization (Section \ref{sec:opt:geo}) takes less than 10 seconds, and the majority rest is for plane, camera pose and texture optimization (Section \ref{sec:opt:plane}). Note that our code is CPU version only and it currently does not contain any parallel acceleration technique in all steps. As we describe in Section \ref{sec:opt:plane}, optimization of each plane is independent with each other and the process can be run in parallel. So is the same for the optimization of each camera pose and correction function in each frame. Moreover, we found that the majority of time in the optimization process is spent on computing the Jacobian matrix in Gaussian Newton algorithm, which can be greatly accelerated using GPU since each texel's computation process is also independent with all the others. 

For comparison, we compare our method with two state-of-the-art systems: BundleFusion \cite{dai2017bundlefusion} and 3DLite \cite{huang20173dlite}. BundleFusion generates a dense and fused reconstruction in real-time, and 3DLite is similar to our method that it takes as input a dense reconstruction from BundleFusion, generates lightweight mesh with face textures by extracting large planar regions as geometry and optimizing texture on it. Figure \ref{fig:comp} shows comparison of mesh with face textures between these methods. BundleFusion models are dense with oversmoothed sharp features and coarse textures. 3DLite models are neat and clear with sharp textures. However, it extracts only large planar regions from the scene as the final mesh, and obviously misses almost all geometry objects on or adhering to planes including both large structured ones (such as book shelf in the 1st row of Figure \ref{fig:comp}) and freeform objects with curved surfaces (such as objects on the table in the 2nd row of Figure \ref{fig:comp}). Our method exceeds them in solving their aforementioned problems. Moreover, 3DLite also generates many misaligned artifacts in the face textures while our method is better in these relevant places (such as the map on the wall in the 3rd row of Figure \ref{fig:comp}).

\begin{table}[!]
\begin{center}
\caption{Quantitative data of RGB-D scans and our results. Here $|V|$ is number of vertices, $|F|$ is the number of faces, $|K|$ is the number of keyframes, $t$ is the total running time of our entire pipeline in seconds. The bottom row `fr3/loh' is shorted for `fr3/long\_office\_household' model.}
\begin{tabular}{|c|c|c|c|c|c|c|}
\hline
Scan & \multicolumn{3}{c|}{Input} & \multicolumn{3}{c|}{Result}\\
\cline{2-7}
 & $|V|$ & $|F|$ & $|K|$ & $|V|$ & $|F|$ & $t$(s) \\
\hline
copyroom & 3.70M & 7.28M & 895 & 55.2K & 104K & 1850 \\
\hline
apt0 & 7.83M & 15.4M & 860 & 84.6K & 160K & 2125 \\
\hline
office0 & 5.71M & 11.3M & 616 & 68.5K & 130K  & 1439 \\
\hline
office1 & 6.03M & 11.9M & 573 & 69.1K & 129K  & 1331 \\
\hline
office2 & 5.63M & 11.0M & 700 & 73.6K & 135K & 1886 \\
\hline
office3  & 6.36M & 12.6M & 763 & 56.7K & 108K & 1972 \\
\hline
of kt2  & 1.20M & 2.36M & 176 & 14.9K & 27.4K & 986 \\
\hline
lr kt2n & 
1.14M & 2.25M & 176 & 22.1K & 41.9K  & 1128 \\
\hline
fr2/desk & 1.37M & 2.69M & 372 & 37.6K & 73.4K  & 787 \\
\hline
fr3/loh & 2.42M & 4.75M & 243 & 43.0K & 83.7K  & 576 \\
\hline
\end{tabular}\label{tables}
\end{center}
\end{table}

\textbf{Limitations}. While our method can generate lightweight textured mesh with sharp features and geometry details preserved, it still contains several limitations as we found in experiments. Firstly, our face textures are not as sharp as 3DLite's since the latter introduces many techniques to optimize texture, such as texture sharpening and color correction across frames. However, these steps are very time-consuming and possibly takes hours, according to the running time data described in the 3DLite paper \cite{huang20173dlite}. We plan to find a similar but faster way to further optimize textures. Moreover, our method cannot fill holes and gaps that always appears in the RGB-D scans, while 3DLite can generate a complete geometry from extracted large planes by extrapolating existing planes and filling holes. However, it is still a challenging problem to complete the geometry of a noisy reconstruction without removing any geometry details. Additionally, our texture optimization process is similar to Zhou and Koltun \cite{zhou2014color} which relies on dense photometric error, and sensitive to initial input during camera pose optimization and easy to end in local minima \cite{huang20173dlite}. Therefore, if large error exists in initial camera poses generated from BundleFusion system, our results may contain misaligned face textures.

\section{Conclusion}

We have presented a novel approach to generate lightweight reconstruction with clear face texture from an initial 3D reconstruction with dense and fused mesh. Unlike existing methods which only detect large planar regions in the scene, our method detects planes from all objects and partitions the entire mesh by planes. Then, we simply the dense mesh based on planar partition, and optimize planes, camera poses and face textures to maximize the photometric consistency across frames, and finally optimize mesh geometry to maximize plane-geometry consistency. Experimental results demonstrate that our reconstruction can preserve sharp features and geometry details in the mesh very well. We believe that our method can be applied in relevant situations acquiring textured indoor scene reconstruction without missing geometry details. In the future, we plan to improve our method by solving limitations aforementioned, and increase the efficiency of our method by introducing GPU acceleration or other related techniques.

\clearpage

{\small
\bibliographystyle{ieee}
\bibliography{egbib}
}

\end{document}